\documentclass[10pt,twocolumn,letterpaper]{article}

\usepackage{cvpr}              

\usepackage{graphicx}
\usepackage{amsmath}
\usepackage{amssymb}
\usepackage{booktabs}
\usepackage[accsupp]{axessibility} 

\usepackage[pagebackref,breaklinks,colorlinks]{hyperref}

\usepackage[capitalize]{cleveref}
\crefname{section}{Sec.}{Secs.}
\Crefname{section}{Section}{Sections}
\Crefname{table}{Table}{Tables}
\crefname{table}{Tab.}{Tabs.}

\def\cvprPaperID{2} 
\def\confName{CVPR}
\def\confYear{2023}

\begin{document}

\title{Motion-state Alignment for Video Semantic Segmentation}

\author{Jinming Su$^*$, Ruihong Yin$^*$, Shuaibin Zhang and Junfeng Luo \\
Meituan \\
{\tt\small \{sujinming, yinruihong, zhangshuaibin02, luojunfeng\}@meituan.com}
}
\maketitle

\begin{abstract}
In recent years, video semantic segmentation has made great progress with advanced deep neural networks. However, there still exist two main challenges \ie, information inconsistency and computation cost. To deal with the two difficulties, we propose a novel motion-state alignment framework for video semantic segmentation to keep both motion and state consistency. In the framework, we first construct a motion alignment branch armed with an efficient decoupled transformer to capture dynamic semantics, guaranteeing region-level temporal consistency. Then, a state alignment branch composed of a stage transformer is designed to enrich feature spaces for the current frame to extract static semantics and achieve pixel-level state consistency. Next, by a semantic assignment mechanism, the region descriptor of each semantic category is gained from dynamic semantics and linked with pixel descriptors from static semantics. Benefiting from the alignment of these two kinds of effective information, the proposed method picks up dynamic and static semantics in a targeted way, so that video semantic regions are consistently segmented to obtain precise locations with low computational complexity. Extensive experiments on Cityscapes and CamVid datasets show that the proposed approach outperforms state-of-the-art methods and validates the effectiveness of the motion-state alignment framework.
\end{abstract}
\let\thefootnote\relax\footnotetext{$^*$ Equal contribution.}

\begin{figure}[t]
\centering
\includegraphics[width=1\columnwidth,height=7.2cm]{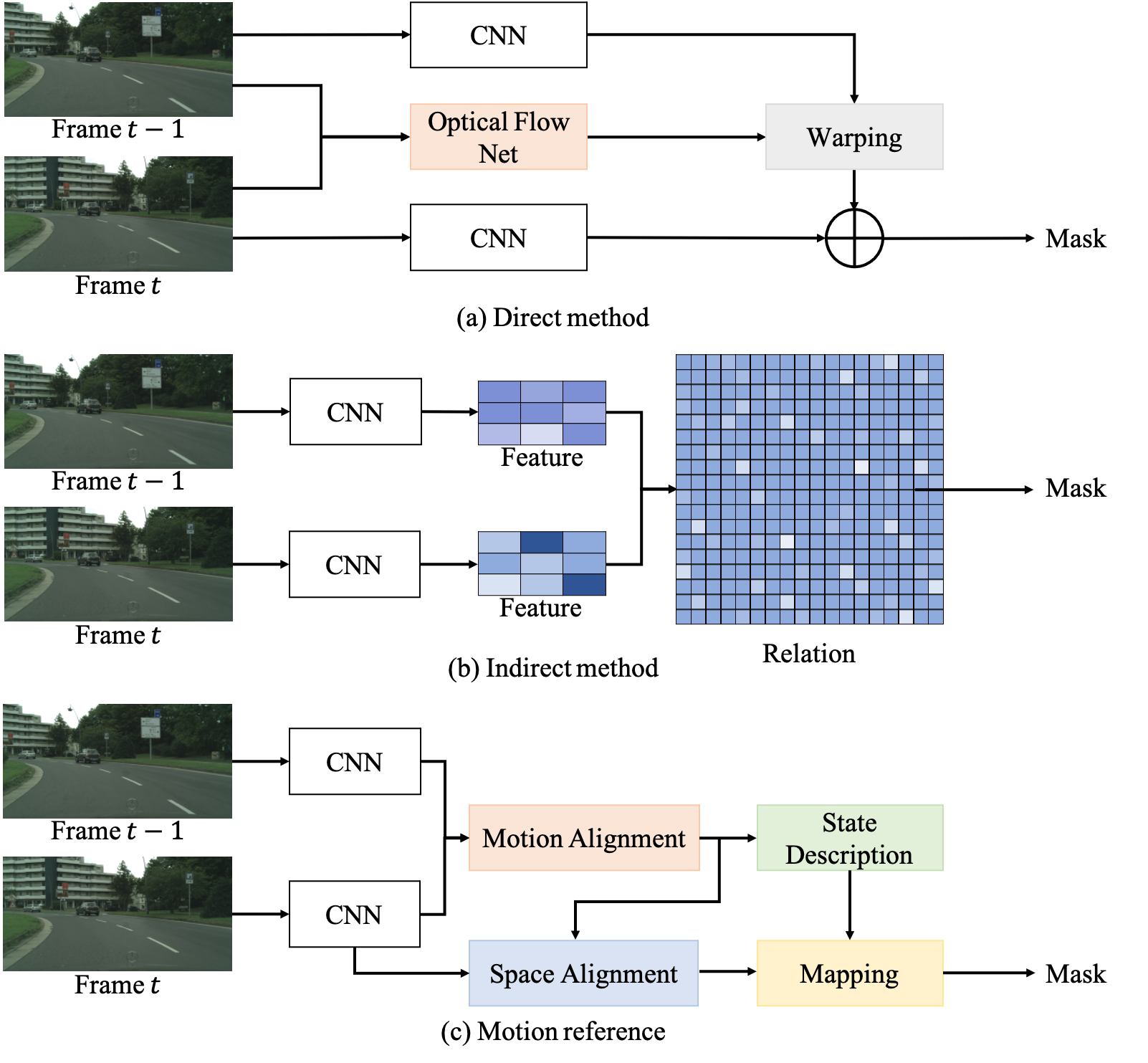}
\vspace{-0.5cm}
\caption{Comparison of different methods for video semantic segmentation. (a) Direct methods explicitly learn the pixel movement depending on the pretrained optical flow network, which leads to information inconsistency. (b) Indirect methods implicitly model the relationship between all pixels with the attention mechanism, which usually results in high computation costs. (c) The proposed motion-state alignment framework decouples the task into motion alignment, state alignment, and semantic assignment, which ensures consistent and efficient learning of semantics.}
\label{fig:motivation}
\end{figure}

\begin{figure*}[t]
\centering
\includegraphics[width=1\textwidth,height=6.1cm]{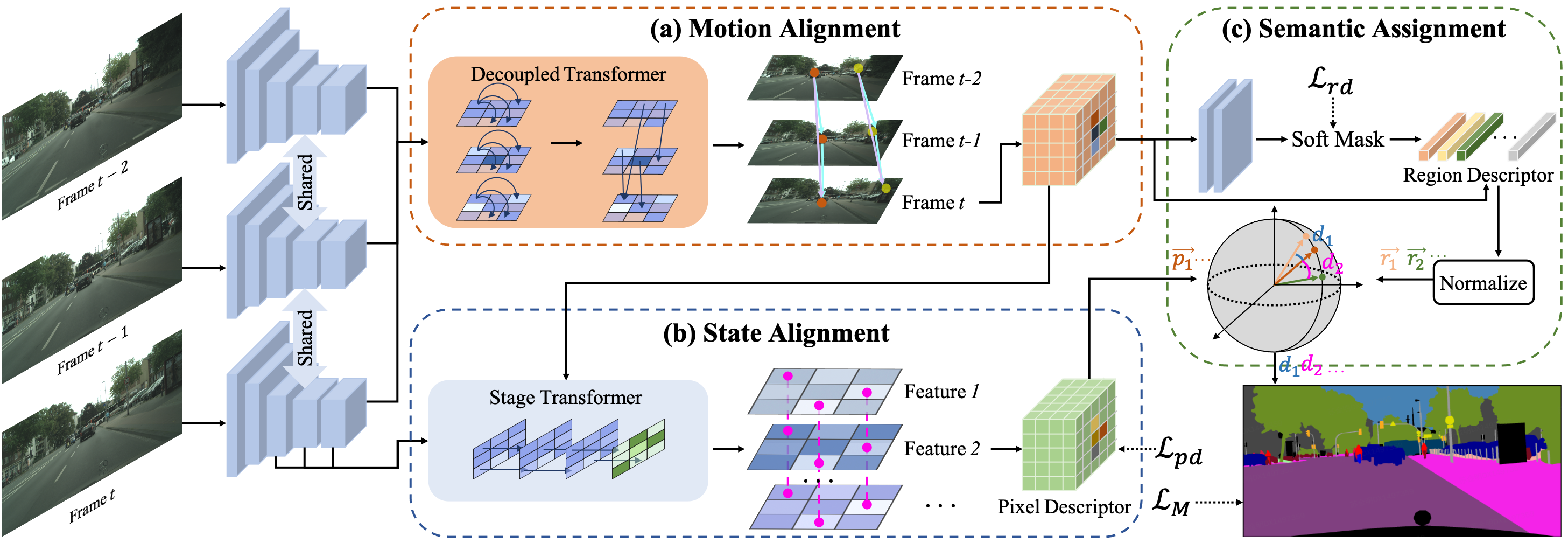}
\vspace{-0.5cm}
\caption{Framework of the proposed motion-state alignment framework. We first extract the independent features for each frame in one video with the shared feature extractor. Then, the dynamic semantics and static semantics about the current frame are aligned via motion and state alignment branches to produce region descriptors and pixel descriptors. Finally, each pixel descriptor is linked to the corresponding region descriptor in a partitioned way to obtain the final segmentation result.}
\label{fig:framework}
\end{figure*} 

\section{Introduction}
\label{sec:intro}

Semantic segmentation \cite{long2015fully,chen2017deeplab,he2019dynamic}, as a dense prediction task, assigns every pixel with the category. Based on image-based semantic segmentation, video semantic segmentation~\cite{zhu2017deep,zhu2019improving,liu2020efficient} introduces temporal information for each frame, which is very challenging and has attracted lots of attention. Recently, video semantic segmentation has made good progress and been widely applied to autonomous driving \cite{hu2020temporally}, video surveillance \cite{jin2017video}, and other fields.

For video semantic segmentation, the key is how to utilize temporal information to improve the accuracy and consistency of segmentation results across frames. Towards this end, various methods based on deep learning \cite{jain2019accel,paul2021local,wang2021temporal} have been proposed to conduct video semantic segmentation recently. From the perspective of feature representation and use, these existing methods can be summarized into two main categories, \ie,
direct methods and indirect methods. Direct methods (as shown in Fig.~\ref{fig:motivation} (a))\cite{zhu2017deep,gadde2017semantic,zhu2019improving} usually use a separately pretrained optical flow network \cite{weinzaepfel2013deepflow,revaud2015epicflow,dosovitskiy2015flownet,ilg2017flownet} to propagate information between adjacent frames without fine-tune on semantic segmentation dataset, owing to the lack of optical flow annotations. However, there is a domain shift between the datasets for flow and datasets for segmentation, so this method leads to warp inaccuracy and information inconsistency. 
Additionally, optical flow estimation has difficulties in handling occlusion and targets moving out of the frame, which is harmful to segmentation quality. 
The errors of optical flow network will propagate to other frames, which is harmful to the segmentation performance. 
Additionally, optical flow estimate has difficulties in handling occlusion. 
To tackle this dilemma, indirect methods (Fig.~\ref{fig:motivation} (b))\cite{wang2018non,liu2020efficient,hu2020temporally,wang2021temporal} propose to calculate a relationship matrix by attention-based mechanism 
to implicitly align features across multiple frames. 
Different from direct methods, indirect methods can implicitly extract temporal features. 
However, these methods calculate the similarity between all pixels across frames, which results in high computation cost and low inference speed. This limits the application on real-world applications, like autonomous driving. 
Therefore, video semantic segmentation still faces two main difficulties, \ie  information inconsistency and computation cost. Information inconsistency is to model temporal consistency between adjacent frames. In video semantic segmentation, applying segmentation network to separate frame will generate inconsistent results, but the relation between frames can increase information and decrease uncertainty during training and inference. Therefore, instead of using one single frame, encoding temporal relation between frames plays an important role in improving performance. Another difficulty is that deep and complex methods based on attention mechanism \cite{vaswani2017attention, wang2018non} have high computation cost, although they have succeed in achieving high accuracy. It is hard to extend these outstanding segmentation methods to real-world applications, like autonomous driving. Therefore, methods with low computation cost are very non-trivial in real-world scenes.




To address the problem of information inconsistency and high computation cost, we put forward a novel motion-state alignment framework for video semantic segmentation as shown in Fig.~\ref{fig:framework}. 
We divide the inter-frame information in one video into dynamic and static features, and construct corresponding motion and state alignment branches, which respond for maintaining the consistency between dynamic and static semantics respectively. 
In the motion alignment branch, an efficient spatial-temporal decoupled Transformer is established to align motion features across adjacent frames, which can maintain the motion consistency between temporal frames with low computations. 
With a stage Transformer, the state alignment branch is constructed to strengthen static features and maintain state consistency. 
Additionally, a semantic assignment mechanism is proposed to fuse both dynamic and static semantics with semantic partition. 
In the end, all pixels are partitioned into different regions with semantic categories as the final segmentation results. In the proposed framework, we construct different modules to align dynamic and static information respectively, which can effectively keep the information consistent across adjacent frames. Experimental results on two challenging video segmentation datasets have shown the effectiveness and efficiency of the proposed  framework. 

Our contributions are as follows: (1) We propose a novel motion-state  alignment framework for video semantic segmentation to address information inconsistency and high computation costs from the new motion-state perspective. (2) Motion alignment mechanism is constructed to align information between temporal sequences and keep motion consistency. State alignment mechanism is built to conduct pixel-to-pixel alignment between different features and keep static consistency. (3) Our method achieves high accuracy on two challenging datasets: 81.5\% mIoU on Cityscapes and 78.8\% on CamVid, which is superior to the state-of-the-art methods in video semantic segmentation. In addition, our lightweight model with ResNet18 can reach 24.39 FPS on the Cityscapes dataset with competitive mIoU 75.8\%.

\section{Related Work}
\label{sec:relatedwork}
In this section, we review the related works that aim to deal with the challenges of video semantic segmentation. 

\subsection{Direct Methods}
In order to realize temporal consistency, direct methods \cite{gadde2017semantic,zhu2017deep,zhu2019improving} adopt an optical flow network to predict per-pixel correspondence between the current frame and other frames. NetWarp~\cite{gadde2017semantic} computes the flow between adjacent frames and warps intermediate features. 
VPLR~\cite{zhu2019improving} leverages an optical flow network and a spatio-temporal transformer recurrent layer to propagate information between frames. EFC~\cite{ding2020every} jointly trains segmentation network and optical flow network. It takes a lot of time to calculate optical flow between every two frames. To reduce computation cost and speed up the inference, some methods adopt the idea of keyframes and share the feature maps of sparse keyframes. For example, DFF~\cite{zhu2017deep} only runs the expensive feature extraction network on keyframes and then warps the feature maps to non-key frames.  Accel~\cite{jain2019accel} propagates high-detail features on a reference branch 
by means of optical flow estimation. Although direct methods are able to capture temporal information successfully, they still cannot be constrained by the disadvantages of optical flow networks.

\subsection{Indirect Methods}
Indirect methods \cite{wang2021temporal,liu2020efficient,hu2020temporally} are another kind of method to gain temporal consistency. They can avoid the use of optical flow networks. TMANet \cite{wang2021temporal} explores long-range temporal dependencies by means of the self-attention mechanism, which computes a heavy relation matrix between each pixel of neighboring frames. ETC~\cite{liu2020efficient} introduces temporal loss and temporal consistency knowledge distillation methods to improve the temporal consistency. TDNet~\cite{hu2020temporally} divides the complex feature network into shallow sub-networks for each frame and performs an attention propagation module to achieve temporal consistency. In indirect methods, although the dimension of feature maps is small, they still compute the high-dimension relation matrix between pixels. Besides, there is a much closer relationship in pixels of consecutive temporal or spatial context.

\subsection{Video Transformer}
In video understanding, like video super-resolution and video classification, it is a new trend to excavate long-sequence dependencies by ways of Transformer architecture \cite{vaswani2017attention,dosovitskiy2020image,liu2021swin,chu2021twins}. TimeSformer \cite{bertasius2021space} adopts Transformer over the space-time volume on the action classification task.
ViViT \cite{arnab2021vivit} develops several factorized Transformer blocks on spatial and temporal dimensions. VSR-Transformer \cite{yang2020learning} uses a spatial-temporal attention layer to exploit spatial-temporal information in video super-resolution. Transformer is very efficient to capture spatial and temporal relations. However, at present, no method keeps eye on applying Transformer to video semantic segmentation.

\section{The Proposed Approach}
\label{sec:ProposedApproach}

In the section, we first introduce the novel motion-state perspective, and then explain each part of the framework.

\subsection{Motion-state Perspective}
To deal with the task of video semantic segmentation, we rethink this task from a novel motion-state perspective, in which semantic information in the video is divided into two parts, namely dynamic semantics and static semantics. Dynamic semantics refers to semantic information that can be easily extracted from motion, which is usually beneficial to semantic region location. Static semantics refers to semantic information that can be easily extracted at each state point during motion and contain more details of the semantic area.
In this perspective, information consistency consists of motion consistency and state consistency.

Based on the motion-state perspective, we propose a novel motion-state alignment framework for video semantic segmentation (named as \textbf{MSAF}) as depicted in Fig.~\ref{fig:framework}, to address the difficulties (\ie, information inconsistency and computation cost). In this framework, we first extract independent features for each frame of one video. Then, through the motion alignment mechanism and state alignment mechanism, the dynamic semantics and static semantics about the current frame are aligned to guarantee information consistency. Next, pixel descriptors from the state alignment are linked to the corresponding region descriptors from the motion alignment in a semantic partition strategy. Thus, all pixels can be assigned to different regions with precise semantic categories and obtain the final segmentation results. In this way, our proposed method can learn both dynamic and static semantics, thus the video semantic regions are consistently segmented to obtain precise locations with low computational complexity. Details of the proposed approach are described as follows.

\subsection{Feature Extractor}
To extract features for each frame of one video, we take ResNet~\cite{he2016deep} as the feature extractor, which removes the last two layers (\ie, classification and global average pooling layers) for the pixel-level prediction task. Feature extractor has five stages for feature encoding, named as $\mathcal{F}_{s}(\pi_s)$ 
with parameters $\pi_s$, where $s=1,2,\dots, 5$ represent the $s$th stage of the feature extractor. 
For convenience, we use $\mathcal{F}_{s}$ to represent $\mathcal{F}_{s}(\pi_s)$.
To obtain larger feature maps, the strides of all convolutional layers in the residual modules $\mathcal{F}_{4}$ and $\mathcal{F}_{5}$ are set to 1. To further enlarge the receptive fields of high-level features, we set the dilation rates~\cite{yu2016multi} to 2 and 4 for convolution layers in $\mathcal{F}_{4}$ and $\mathcal{F}_{5}$, respectively. Therefore, after the feature extractor, for a $H \times W$ input image, we can get a feature map with the size of $\frac{H}{8} \times \frac{W}{8}$ by the feature extractor. Note that we use the same shared feature extractor to capture features for every frame.

\subsection{Motion Alignment}
\begin{figure}[t]
\centering
\includegraphics[width=1.0\columnwidth]{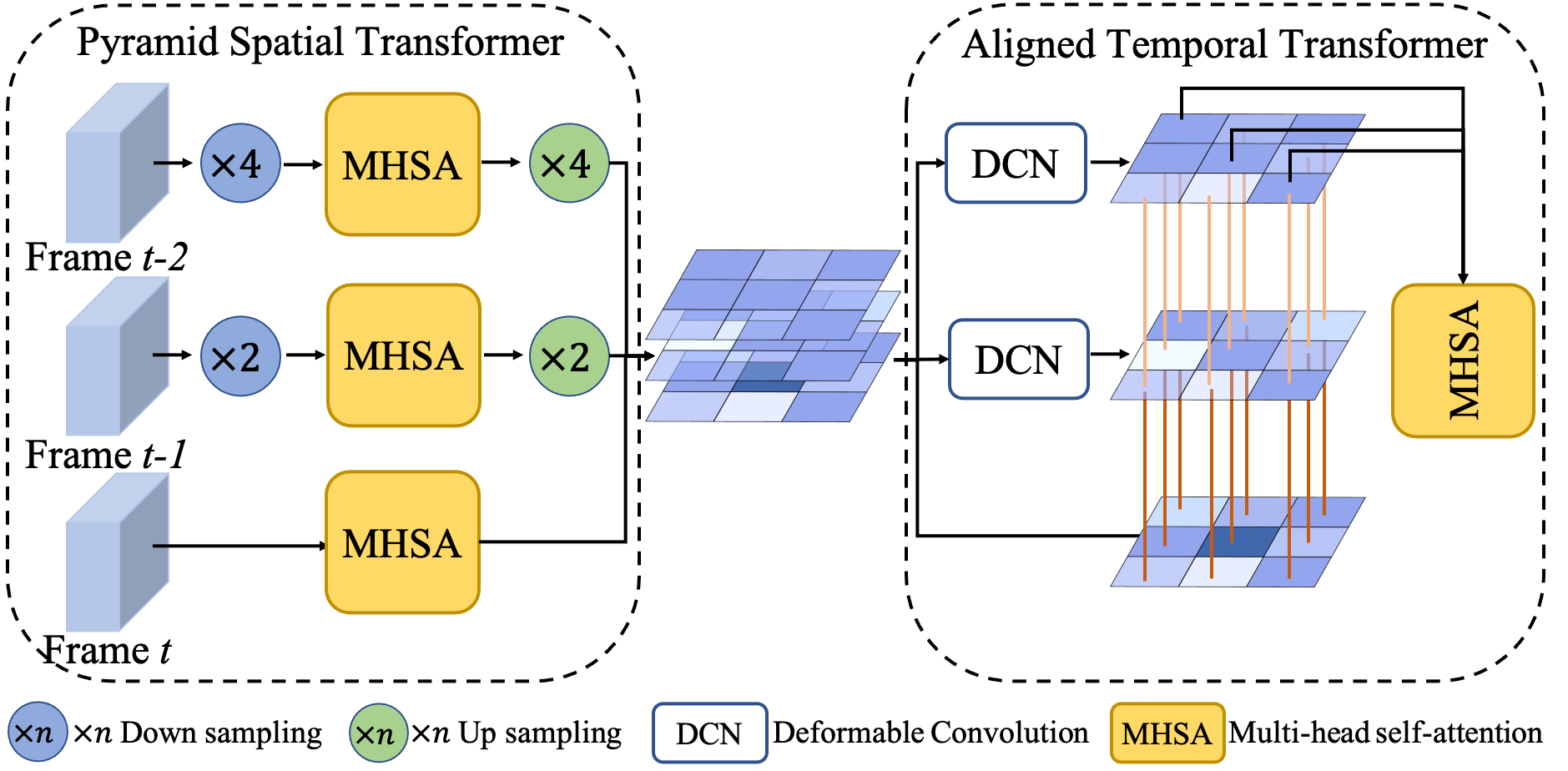}
\vspace{-0.5cm}
\caption{Structure of decoupled Transformer.}
\label{fig:transformer}
\end{figure}

In one video, there are two important benefits of frame information. Firstly, information across adjacent frames can increase perception and decrease uncertainty. Secondly, the information can make it easy to capture the motion regions, which is beneficial to semantic region location.

In order to deal with the information inconsistency in many existing methods, we propose a motion alignment mechanism to capture the motion information in the video and maintain the motion consistency, as shown in Fig.~\ref{fig:framework}. 
Specifically, in order to model the motion consistency between frames, with reference to optical flow that models a learnable ``flow'' to match pixels between frames, we propose an efficient decoupled Transformer to learn the ``flow'' as depicted in Fig.~\ref{fig:transformer}.

\noindent \textbf{Decoupled Transformer.} Transformer \cite{vaswani2017attention} is beneficial to long-term and global dependence, which is the main advantage compared with the ordinary convolutional operation. It is a promising choice to model the long-distance relationship in both time and space dimensions by utilizing Transformer, but it has the disadvantage in a lot of computation. Specifically, we represent the extracted features as $\mathcal{F}^{t}_{5}$ with the size of $\mathcal{C}_{\mathcal{F}_{5}} \times H_{\mathcal{F}_{5}} \times W_{\mathcal{F}_{5}}$, where $t, \mathcal{C}_{\mathcal{F}_{5}}, H_{\mathcal{F}_{5}}$ and $W_{\mathcal{F}_{5}}$  means the frame index, number of the channel, width and height of the feature map. In this paper, we empirically choose three frames with the feature $\mathcal{F}^{t-2}_{5}$, $\mathcal{F}^{t-1}_{5}$ and $\mathcal{F}^{t}_{5}$ as an example. Given a naive explanation, ignoring the effect of scaled dot-product and multi-head on reducing the amount of computation, Transformer can model the relationship between all pixels in a feature map. Thus, when Transformer is utilized to model spatio-temporal information, 
the computation cost of Transformer is as follows:
\begin{equation}
\begin{aligned}
C(\text{Transformer}) = 3n \times n \times n = 3n^3, 
\end{aligned}
\label{equ:transformer}
\end{equation}
where $n = \mathcal{C}_{\mathcal{F}_{5}} \times H_{\mathcal{F}_{5}} \times W_{\mathcal{F}_{5}}$ means the number of pixels in a feature map. Considering frames $t-2$, $t-1$, and $t$, the computation cost that constructs the relation between one pixel with all other pixels in spatio-temporal dimension is $n \times n$. Therefore, the total computation cost between frame $t$ and $t-1$ as well as $t-2$ is $n^3$. Therefore, the computation cost of vanilla Transformer is $3n^3$. Usually, this computation cost is tolerable, unless the size of feature graph is relatively large.

To reduce the computation cost, we consider decoupling the spatio-temporal relationship and propose a decoupled Transformer (D-Transformer) as shown in Fig.~\ref{fig:transformer}. D-transformer consists of two modules \ie, pyramid spatial transformer (PST) and aligned temporal transformer (ATT). 

In PST, features captured by the feature extractor for each frame are as the input. Next, features are downsampled to reduce the amount of computation. The distant frames in the temporal dimension provide more abstract information with a larger sampling rate ($\times 4$ in frame $t-2$), while the near frames provide more specific information ($\times 2$ in frame $t-1$). Then, features after sampling are fed to the multi-head self-attention module (MHSA)~\cite{vaswani2017attention} to model global dependency in the spatial dimension. After that, the corresponding upsampling operation is adopted to ensure the precise alignment in spatial dimensions. PST block is formally defined as:
\begin{equation}
\begin{aligned}
& \mathcal{D}^{t-2}_{\text{PST}} = \text{US}^{t-2}(\text{MHSA}^{t-2}_{spa}(\text{DS}^{t-2}(\mathcal{F}^{t-2}_{5}))) \\
& \mathcal{D}^{t-1}_{\text{PST}} = \text{US}^{t-1}(\text{MHSA}^{t-1}_{spa}(\text{DS}^{t-1}(\mathcal{F}^{t-1}_{5}))) \\
& \mathcal{D}^{t}_{\text{PST}} = \text{MHSA}^{t}_{spa}(\mathcal{F}^{t}_{5})
\end{aligned},
\end{equation}
where $\text{US}$ and $\text{DS}$ mean up-sampling and down-sampling operation. $\text{MHSA}_{spa}$ means multi-head self-attention module in the spatial dimension with the following formulation:
\begin{equation}
\begin{aligned}
& X' = \text{Attention}(\text{LN}(X)) + X \\
& \text{MHSA}_{spa} = \text{MLP}(\text{LN}(X')) + X'
\end{aligned},
\label{eq:preception_loss}
\end{equation}
where $X$ is the input tensor, LN is layer normalization layer ~\cite{ba2016layer}, MLP is multi-layer perceptron, and Attention is the self-attention module referring to~\cite{vaswani2017attention} for details. In this way, spatial features of neighboring frames are extracted  in a pyramid way, and the computation cost is as follows (note  we ignore the reduction of computation caused by downsampling for fair):
\begin{equation}
\begin{aligned}
C(\text{PST}) = 3 \times n^2 = 3n^2. 
\end{aligned}
\end{equation}

In ATT, three feature maps of different frames after PST are used as input. Due to motion, a pixel at the current frame may not match the pixel at the same location of adjacent frames. Therefore, it is necessary to first align the spatial information pixel by pixel in the temporal dimension. For this purpose, we introduce deformable convolutional layers ~\cite{dai2017deformable} in adjacent frames to extract the offset of each pixel caused by motion. The detail is as follows:
\begin{equation}
\begin{aligned}
Y(p) = \sum^{K \times K}_{k=1} w_{k} \cdot X(p + p_k + \Delta p_k), 
\end{aligned}
\end{equation}
where $K$ defines the number of points that may correspond to the pixel at this position in the adjacent frame, $X(p)$ is the pixel at location $p$ of input feature  $X$, and $Y(p)$ represents the pixel at location $p$ of input feature  $Y$. $p_{k}$ means the location range of $K \times K$ centered on the current position, $w_k$ is the weight value and $\Delta p_k$ is the learned offset.
In this way, each pixel can learn the position of the associated pixel in adjacent frames. Here we set $K=3$ to perceive multiple positions and ensure the information alignment between frames. Then, each pixel at the current frame and its associated pixels in adjacent frames are passed together to MHSA to model the temporal long-term dependency in the spatial dimension. The ATT block is formally defined as:
\begin{equation}
\mathcal{D}_{\text{ATT}} = \text{MHSA}_{tem}(\text{DCN}^{t-2}(\mathcal{D}^{t-2}_{\text{PST}}), \text{DCN}^{t-1}(\mathcal{D}^{t-1}_{\text{PST}}), \mathcal{D}^{t}_{\text{PST}}),
\label{eq:preception_loss}
\end{equation}
where DCN is the deformable convolutional operation and $\text{MHSA}_{tem}$ means multi-head self-attention in the temporal dimension. DCN is only used in neighbouring two frames to align a pixel in the current frame with pixel in neighbouring frames. In this way, we can get the computation cost of ATT as follows (note the computation cost of DCN can be regarded as $K^2n$):
\begin{equation}
\begin{aligned}
C(\text{ATT}) = 2 \times 3n + 2 \times 9n = 24n. 
\end{aligned}
\end{equation}

Therefore, the whole computation cost of D-transfomer is:
\begin{equation}
\begin{aligned}
C(\text{D-Transformer}) &= C(\text{PST}) + C(\text{ATT}) \\
                        &= 3n^2 + 24n
\end{aligned}.
\label{equ:d-transformer}
\end{equation}
Comparing Equation (\ref{equ:transformer}) with (\ref{equ:d-transformer}), we can clearly see that the value of C(\text{Transformer}) - C(\text{D-Transformer}) increases monotonously and linearly in the range $[\frac{1 + \sqrt{33}}{2}, +\infty)$, and $\lim_{n \to +\infty} \frac{C(\text{Transformer})}{C(\text{D-Transformer})} = n$. Obviously, $n \gg \frac{1 + \sqrt{33}}{2}$, which means that the computation cost of D-Transformer is much smaller than original Transformer and D-Transformer is much more efficient.

Therefore, D-Transformer can construct the spatio-temporal relationship between video frames, and implicitly ensure the alignment of motion information between frames, which is beneficial to ensure motion consistency. 

For convenience, motion alignment branch is denoted as $\mathcal{M}(\pi_{\mathcal{M}})$ with parameters $\pi_{\mathcal{M}}$. The output of $\mathcal{M}$ including three features \ie, $\mathcal{M}^{t-2}$, $\mathcal{M}^{t-1}$ and $\mathcal{M}^{t}$ that all have the same shape with $\mathcal{F}^t_{5}$, which captures the dynamic information of one video. In order to obtain semantics, $\mathcal{M}$ is supervised by the ground-truth semantic mask, and the supervision signal will be mentioned later. Therefore, $\mathcal{M}$ carries rich dynamic semantics. In this manner, motion alignment efficiently maintains motion consistency.

\subsection{State Alignment}
Except for dynamic information, static information also plays an important role in video semantic segmentation. The static semantics can be easily extracted from a frame described as a motion state at a certain moment, which is usually more conducive to restoring the details of semantic areas. In order to ensure the expressive ability of state information in feature space, we align the state information of different feature sources and propose a state alignment mechanism as depicted in Fig.~\ref{fig:framework}.

\noindent \textbf{Stage Transformer.} As neural network goes deeper, low-level features contain more details of images, and high-level features have more semantic information. To contain both detail and semantic information, in the state alignment branch, we collect features at different levels of the current frame (regarded as state features) and feed them to the proposed stage Transformer (S-Transformer), which can gain detailed information and model the relationship of corresponding pixels between features of different levels in the stage dimension. The S-Transformer block is formally defined as:
\begin{equation}
\mathcal{S}_{TR} = \text{MHSA}_{sta}(\mathcal{F}^t_{3}, \mathcal{F}^t_{4}, \mathcal{F}^t_{5}, \mathcal{M}^t),
\end{equation}
where $\text{MHSA}_{sta}$ is the multi-head self-attention module in the stage dimension. Moreover, to enhance the features of current frames, we also add the features $\mathcal{M}^t$ of current frames from motion alignment to S-Transformer. In this way, static information from different stage and dynamic information of current frame are aligned together and modeled for the state consistency.

\noindent \textbf{Pixel Descriptor.} We define the state alignment branch as $\mathcal{S}(\pi_{\mathcal{S}})$ with parameters $\pi_{\mathcal{S}}$, 
The output of $\mathcal{S}$ then goes through two convolutional layers to gain pixel descriptor. For convenience, we name the pixel descriptor as $\mathcal{P}$. To obtain the semantic information, $\mathcal{P}$ is supervised by the ground-truth mask of semantic segmentation (represented as $G$) by minimizing the loss:
\begin{equation}
\mathcal{L}_{pd} = CE(\mathcal{P}, G),
\label{equ:l_pd}
\end{equation}
where $CE(\cdot, \cdot)$ is the cross-entropy loss function with the following formulation:
\begin{equation}
CE(P, G) = -\sum^{H \times W}_{i=1} \sum^{C}_{j=1} G_{i,j}\mathrm{log}P_{i,j},
\end{equation}
where $P_{i,j}$ and $G_{i,j}$ represents the predicted and real probabilities that the $i$th ($i \in \mathbb{R}^{H \times W}$) pixel belongs to $j$th ($1 \le j \le C$) class, respectively. Under the supervision, we regarded each pixel $pix \in \mathbb{R}^{\mathcal{C}_{\mathcal{F}_{5}} \times H_{\mathcal{F}_{5}} \times W_{\mathcal{F}_{5}}}$ in $\mathcal{P}$ as the pixel descriptor, which represents the static semantics and state details of each pixel.

\subsection{Semantic Assignment}
After obtaining the motion features $\mathcal{M}^t$ and state features $\mathcal{P}$, we propose the semantic assignment mechanism to produce the final segmentation results as shown in Fig.~\ref{fig:framework}.

\noindent \textbf{Region Descriptor.} From motion information, it is easy to extract region-level information, which is beneficial to semantic region location. Therefore, we introduce the region descriptors to characterize the region-level features.

To generate region descriptors, we first add two convolutional layers after $\mathcal{M}^t$ to obtain dynamic information. After the two convolutional layers, the supervision from the ground-truth semantic mask $G$ is imposed to produce the soft mask $\mathcal{M}^{t'}$ by minimizing the loss:
\begin{equation}
\mathcal{L}_{rd} = CE(\mathcal{M}^{t'}, G).
\label{equ:l_rd}
\end{equation}
Then, we weight $\mathcal{M}$ with each logit (feature map before softmax) of $\mathcal{M}^t$, and average the result by global pooling in the spatial dimension to obtain region descriptors $\mathcal{R} \in \mathbb{R}^{Cls \times 1 \times 1}$ with $Cls$ is the number of semantic categories. Each region descriptor $r \in \mathcal{R}$ represents the region of one semantic category.

\noindent \textbf{Semantic Partition.} With region descriptors and pixel descriptors with similar semantics, we propose the semantic partition strategy to link pixel descriptors to corresponding region descriptors as shown in Fig.~\ref{fig:framework}.

In order to judge the association between pixels and regions, we use the minimum cosine distance as the evaluation metric. Specifically, we compute all the distance between pixel descriptor $pix_{i} \in \mathcal{P}, i=1,2,\dots,H_{\mathcal{F}_{5}} \times W_{\mathcal{F}_{5}}$ with each region descriptor $r_{j} \in \mathcal{R}, j=1,2,\dots, Cls$, and then the index with the minimum distance is taken as the semantic category. This processing is formulated as:
\begin{equation}
\begin{aligned}
CLASS_{pix_{i}} &= \text{argmin}_j(d_1, d_2, \dots, d_j, \dots, d_{Cls}) \\
d_j &= 1 - \cos(pix_{i}, r_j) \\
\cos(pix_{i}, r_j) &= \frac{pix_{i} \cdot r_j}{|pix_{i}||r_j|}
\end{aligned},
\label{equ:class}
\end{equation}
where $|\cdot|$ means the magnitude of a vector. In Equation (\ref{equ:class}), $pix_{i} \cdot r_j$ is dot product of $pix_{i}$ and $r_j$. For each $d_j$, the ignoring magnitude operation about $|pix_{i}|$ doesn't affect the loss value because of the same $pix_{i}$, and we add normalization operation to $r_j$ for each region descriptors. In this way,  Equation (\ref{equ:class}) can be reduced to a matrix multiplication (\ie, the matrix multiplication of $\mathcal{P}$ and $\mathcal{R}$) followed by an argmin operation in channel dimension. After this, we obtain the final results $M$, which is expected to approximate the ground-truth masks $M$ by minimizing the loss:
\begin{equation}
\mathcal{L}_{M} = CE(M, G),
\label{equ:l_M}
\end{equation}
For convenience, semantic assignment branch is denoted as $\mathcal{A}(\pi_{\mathcal{A}})$ with parameters $\pi_{\mathcal{A}}$.

By taking the losses of Eqs.(\ref{equ:l_pd}), (\ref{equ:l_rd}), and (\ref{equ:l_M}), the overall learning objective can be formulated as follows:
\begin{equation}
\min_{\mathbb{P}} \mathcal{L}_{pd} + \mathcal{L}_{rd} + \mathcal{L}_{M},
\label{equ:l_total}
\end{equation}
where $\mathbb{P}$ is the set of $\{\pi_s\}^5_{s=1}$, $\pi_\mathcal{M}$, $\pi_\mathcal{S}$, and $\pi_\mathcal{A}$.

\section{Experiments and Results}
\label{sec:Experiments}

\subsection{Experimental Setup}
\noindent\textbf{Datasets.} To verify the effectiveness of our proposed method, we evaluate results on two public video semantic segmentation datasets, \ie, Cityscapes~\cite{cordts2016cityscapes} and CamVid~\cite{brostow2008segmentation} datasets. The Cityscapes dataset consists of a large set of video frames from 50 European cities. The dataset provides 5,000 fine annotations (\ie, 2,975 for training, 500 for validation, 1,525 for testing) 
and 20,000 coarse annotations. In Cityscapes, there are 19 semantic categories, and the size of images is 1024x2048. In our experiments, we only use 5,000 finely-annotated images for training and inference. 
In addition, CamVid has 4 videos with 11-class annotations, which are divided into 367/101/233 for training/validation/testing respectively. 
The image resolution in CamVid is 640x960.

\begin{table}[t]
\small
\centering
\caption{Performance of state-of-the-art models on Cityscapes. ``-V'' (``-T'') means on the validation (test) dataset, and a single ``-'' indicates that data cannot be obtained. Note that all our results are in single-scale testing, and all our models are only trained on the training dataset (fine). The best result are in \textbf{\color{red}{\underline{{red}}}}.}
\setlength{\tabcolsep}{0.9mm}{
\renewcommand\arraystretch{0.95}
\begin{tabular}{c | c | c |c c c c c}
\hline
Models & Year & Backbone & mIoU-V & mIoU-T  & FPS & TC \\
\hline
PSPNet18 & 2017 & PSPNet18 & 75.5 & - & 10.99 & 68.5   \\
Accel & 2019 & ResNet18 & - & 72.1 & 2.27 & 70.3 \\
ETC& 2020 & PSPNet18 & - & 73.1 & 9.5 & 70.6\\
TDNet & 2020 & PSPNet18 & 76.8 & - & 11.76 & -\\
STT & 2021 & ResNet18 & 77.3 & -  & 11.5 & 73.0 \\
PC & 2022 & HRNet-w18 & 76.4 & - & - & 71.2 \\
\textbf{Ours} & - & ResNet18$_{512}$ & 75.8 & - &24.39 & 72.2\\
\textbf{Ours} & - & ResNet18$_{768}$ & 77.8 & - &13.54 & 73.1 \\
\textbf{Ours} & - & ResNet18 & 78.2 & 78.3 &8.76 & 73.8\\
\textbf{Ours} & - & PSPNet18 & \textbf{\color{red}{\underline{{78.3}}}} & \textbf{\color{red}{\underline{{78.5}}}} &7.91 & 73.9 \\
\hline
PSPNet50& 2017 & PSPNet50 & 78.1 & - & 4.2 & -  \\
Accel & 2019 & ResNet50 & - & 74.2 & 1.49 & -  \\
TDNet & 2020 & PSPNet50 & 79.9 & 79.4 & 5.62 & -  \\
TMANet& 2021 & ResNet50 & 80.3 & - & 2.14  & -  \\
\textbf{Ours} & - & ResNet50 & 80.4 & \textbf{\color{red}{\underline{{79.8}}}} &2.57 & 73.9  \\
\textbf{Ours} & - & PSPNet50 & \textbf{\color{red}{\underline{{80.7}}}} & - 
&2.37 & 74.1 \\
\hline
PSPNet101 & 2017 & ResNet101 & 79.7 & 78.4 & 2.78 & 69.7  \\
DFF & 2017 & ResNet101 & 68.7& - & 5.59 & 71.4 \\
PEARL & 2017 & ResNet101 & 76.5 & 76.5 & - & -   \\
NetWarp& 2017 & PSPNet101 & 80.6 & - & 0.33 & -   \\
LVS & 2018 & ResNet101 & 76.8 & - & 2.63 & -   \\
GRFP & 2018 & PSPNet101 & 80.2 & -  & -   & -   \\
Accel & 2019 & ResNet101 & - & 75.5 & 1.15  & -   \\
ETC & 2020 & PSPNet101 & 79.5 & - & - & -   \\
LMA  & 2021 & PSPNet101 & 78.5 & - &1.32 & -   \\
PC & 2022 & ResNet101 & 76.3 & - &- & 72.4  \\
\textbf{Ours} & - & ResNet101 & 81.3 & 80.0&1.96 & 74.8 \\
\textbf{Ours} & - & PSPNet101 & 81.5 & 81.1   &1.82 & 75.2 \\
\textbf{Ours} & - & HRNetW48 & \textbf{\color{red}{\underline{{82.1}}}} & \textbf{\color{red}{\underline{{81.5}}}}  
& 2.02 & 75.4 \\
\hline
\end{tabular}
}
\label{tab:performance_cityscapes}
\end{table}

\noindent \textbf{Evaluation Metrics.} 
Our method is evaluated on the accuracy, efficiency, and temporal consistency. 
We apply mean Interaction over Union(mIoU) to report the accuracy. mIoU averages the IoU value in all valid semantic categories of the dataset. \
Frames per second (FPS) is used to present the efficiency of methods. And temporal consistency(TC) \cite{lai2018learning} is adopted to evaluate the temporal stability in video tasks, which calculates the mean flow warping error between two neighboring frames.

\begin{table}[t]
\small
\centering
\caption{Performance of state-of-the-art models on CamVid. ``-V'' means on validation dataset and ``-T'' means on test dataset. The best result are in \textbf{\color{red}{\underline{{red}}}}.}
\setlength{\tabcolsep}{2mm}{
\renewcommand\arraystretch{0.95}
\begin{tabular}{c | c | c |c c c }
\hline
Models & Year & Backbone & mIoU-V & mIoU-T  & FPS \\
\hline
MC & 2018 & SegNet & - & 63.7 & 0.5 \\
Accel & 2019 & ResNet18 & - &66.7 &5.88  \\
TDNet & 2020 & PSPNet18 & - & 72.6 & 25.00 \\
STT & 2021 & PSPNet18 & 76.1 & - & 24.7 \\
PC & 2022 & HRNetW18 & 73.2 & - & - \\
\textbf{Ours} & - & ResNet18 & \textbf{\color{red}{\underline{{76.6}}}} & \textbf{\color{red}{\underline{{75.7}}}} & 25.13 \\
\hline
NetWarp & 2017 & Dilation & - & 67.1 &2.53  \\
GRFP & 2018 & Dilation8 & - &66.1 &4.35   \\
Accel & 2019 & ResNet50 & - &67.7 & 4.18  \\
TDNet & 2020 & PSPNet50 & - & 76.0 & 11.11 \\
EFC & 2020 & Dilation8 & - &67.4 & -   \\
TMANet & 2021 & ResNet50 & 76.5 & - & 7.72  \\
\textbf{Ours} & - & ResNet50 & \textbf{\color{red}{\underline{{77.4}}}} & \textbf{\color{red}{\underline{{77.2}}}}  &9.26 \\
\hline
Accel & 2019 & ResNet101 & - &69.3 & 3.13 \\
PC & 2022 & ResNet101 & 75.2 & - & - \\
\textbf{Ours} & - & ResNet101 &77.6 & 77.5& 6.54 \\
\textbf{Ours} & - & PSPNet101 & \textbf{\color{red}{\underline{{77.9}}}} & 78.0& 5.99\\
\textbf{Ours} & - & HRNetW48 &77.7 & \textbf{\color{red}{\underline{{78.8}}}} & 6.14  \\
\hline
\end{tabular}
}
\label{tab:performance_camvid}
\end{table}


\noindent \textbf{Training and Inference.} We verify the robustness of our methods on a series of backbones, \ie, ResNet18 \cite{he2016deep}, ResNet50, ResNet101, PSPNet18 \cite{zhao2017pyramid}, PSPNet50, PSPNet101, HRNet~\cite{wang2020deep}. During training, our method is optimized by AdamW \cite{loshchilov2017decoupled} with a batch size of 2 per GPU. We apply a ``poly'' learning rate policy, and the initial learning rate is 0.00006. A linear warm-up strategy is used in the first 1,500 iterations. We train the model for 80,000 iterations in total. The images are randomly cropped by 769x769 on Cityscapes and  640x640 on CamVid. Data augmentation, like flip and resize, is adopted in our method.

\subsection{Comparisons with state-of-the-art methods}
Our method is compared with state-of-the-art methods including  
PEARL~\cite{jin2017video}, NetWarp~\cite{gadde2017semantic}, LMA~\cite{paul2021local}, DFF~\cite{zhu2017deep}, GRFP~\cite{nilsson2018semantic}, TMANet~\cite{wang2021temporal}, DFF~\cite{zhu2017deep}, LVS~\cite{li2018low},  MC~\cite{huang2018efficient}, Accel~\cite{jain2019accel}, ETC~\cite{liu2020efficient}, TDNet~\cite{hu2020temporally}, EFC~\cite{ding2020every}, STT~\cite{li2021video} and PC~\cite{zhang2022perceptual}.

In Table~\ref{tab:performance_cityscapes}, we compare our method with recent state-of-the-art methods on the Cityscapes dataset. Our method shows great improvement in accuracy on multiple backbones, which shows the robustness of our method. Our method based on ResNet18 has made great achievement with mIoU 78.5\% on the test set, which outperforms other lightweight methods on both val and test sets. Additionally, our ResNet18-based method with input size 512x1024 reaches 24.39 FPS, which can achieve real-time inference. With backbone ResNet50 and PSPNet50, our method also shows great accuracy and speed. In addition, on complex backbones, our method achieves a mIoU of 80.0\%, 81.1\%, 81.5\% with backbone of ResNet101, HRNet48, PSPNet101 respectively. Our method with a complex backbone also has advantages in inference time.


\begin{figure*}[t]
\centering
\includegraphics[width=1\textwidth]{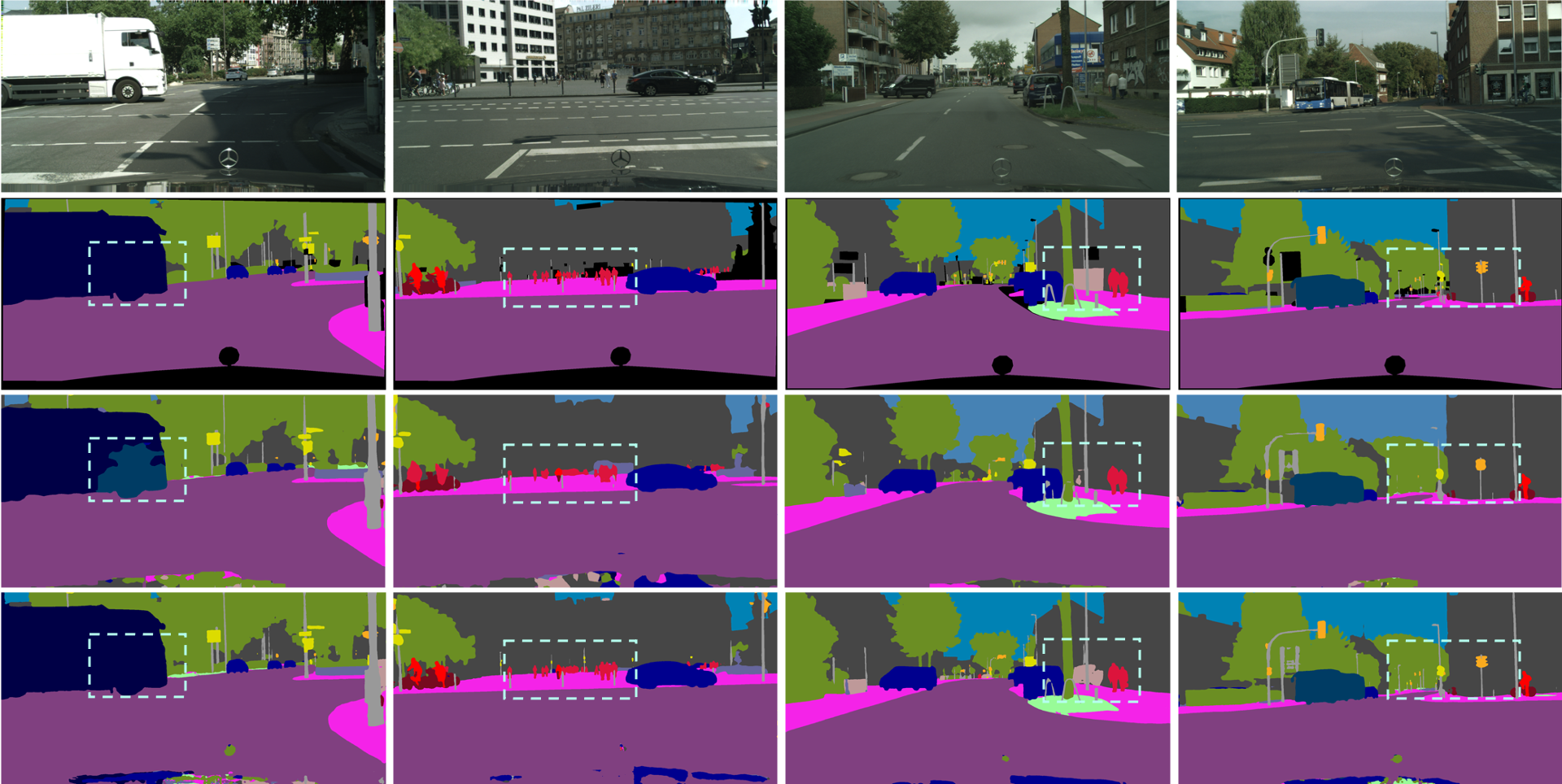}
\vspace{-0.5cm}
\caption{Qualitative results on Cityscapes validation set. From top to bottom: original images, the ground truth, LMA, the segmentation results of our ResNet101 based model.}
\label{fig:visualize}
\end{figure*} 

As listed in Table~\ref{tab:performance_camvid}, we also validate the robustness of our method on the CamVid dataset. We can see that our method with different backbones (\ie, ResNet18, ResNet50, ResNet101, PSPNet101) has achieved significant improvement on both accuracy and efficiency. In particular, compared with Accel in different backbones, our method has significantly improved and runs 2-4 times faster.

In Table~\ref{tab:performance_cityscapes}, we also compare Temporal Consistency (TC) with state-of-the-art methods. The TC of our MSAF is higher than other methods, which verifies that our MSAF can achieve alignment between frames implicitly without the help of optical flow algorithms.

In Fig.~\ref{fig:visualize}, we compare the qualitative results with the state-of-the-art method. We can see that our proposed method can segment the images much more precisely in some complex scenes. These show that our MSAF can extract better semantics information for the segmentation task.

\subsection{Ablation Analysis}

In Table~\ref{tab:performance_ablation}, we carry out ablation experiments on the Cityscapes validation set with backbone ResNet50 to present the effectiveness of each part in our method. 

To investigate the effectiveness of the motion alignment, we replace the motion alignment with a vanilla Transformer (named baseline+TR in Table~\ref{tab:performance_ablation}), which has more parameters than the whole framework MSAF. Although baseline+TR has shown an increase of 3.0\% compared with baseline (74.5\%), our method only with motion alignment (Motion only) achieves better accuracy (79.1\%). This explains that our improvement is not due to the use of the Transformer module, but attributed to the design of motion alignment with decoupled transformer for video semantic segmentation. To provide evidence for this further, we remove the motion alignment (w/o Motion) in our MSAF. In this way, there is no temporal information in ``w/o Motion". Compared with w/o Motion, MSAF gains an improvement by mIoU 3.1\%, which shows that motion alignment is helpful to guarantee the temporal consistency in video semantic segmentation. By the way, DCN brings a gain of 0.6\%.


\begin{table}[t]
\small
\centering
\caption{Performance of different settings of the proposed method (ResNet-50 based) on Cityscapes validation dataset.}
\setlength{\tabcolsep}{3.5mm}{
\renewcommand\arraystretch{0.95}
\begin{tabular}{c | c c c | c}
\hline
& MonA & StaA & SemA & mIoU\\
\hline
Baseline & & & & 74.5\\
Baseline + TR & & & &77.5 \\
Motion only & \checkmark & & & 79.1 \\
State only & & \checkmark & & 76.5\\
w/o Motion & & \checkmark & \checkmark  &77.3\\
w/o State & \checkmark & & \checkmark & 79.7\\
w/o Semantics & \checkmark & \checkmark & & 79.8 \\
\hline
w/o DCN & \checkmark & \checkmark & \checkmark & 79.8\\
MSAF & \checkmark & \checkmark & \checkmark & \textbf{\color{red}{\underline{{80.4}}}} \\
\hline
\end{tabular}
}
\label{tab:performance_ablation}
\end{table}

We also conduct an experiment with state alignment only (State only) and without state alignment (w/o State) to validate the effectiveness of state alignment. The feature maps from motion alignment and backbone are concatenated and forwarded to semantic alignment. In Table~\ref{tab:performance_ablation}, we can see state alignment can improve the accuracy from 79.7\% to 80.4\%. This also shows that static information is important for video semantic segmentation. 

To validate the effectiveness of semantic assignment, we remove the semantic assignment (w/o Semantics) and average the masks of motion alignment and state alignment as the final result. We find that the method with semantic assignment achieves 0.6\% improvement.

\section{Conclusion}
\label{sec:Conclusion}
In this paper, we rethink consistency in video semantic segmentation from the perspective of motion and state semantics, and propose a motion-state alignment framework to achieve both dynamic and static consistency. In this framework, we build a motion alignment branch to encode temporal relations with dynamic semantics. State alignment branch is proposed to enhance features at the pixel level to obtain rich static semantics. Finally, semantic assignment links pixel descriptors from state alignment with region descriptors from motion alignment. Experiments on challenging datasets show that our approach outperforms recent state-of-the-art methods in both accuracy and efficiency.

{\small
\bibliographystyle{ieee_fullname}
\bibliography{egbib}
}

\end{document}